\documentclass[num-refs, twocolumn]{wiley-article}

\usepackage{amssymb}  
\usepackage{url}  
\usepackage{setspace}  
\usepackage{float}  
\usepackage{tikz}  

\usepackage[pagebackref,breaklinks,colorlinks]{hyperref}
\usepackage{caption}

\usepackage{multirow}
\usepackage{siunitx}
\DeclareSIUnit\px{px}
\usepackage{pifont}
\usepackage{hhline}
\usepackage{colortbl}
\usepackage[table]{xcolor}

\usepackage{gradientframe}

\def\secref#1{Sec.~\ref{#1}}
\def\figref#1{Fig.~\ref{#1}}

\def\eqref#1{Eq.~(\ref{#1})}

\papertype{Research Article}

\paperfield{Preprint Version}

\title{Reliability Scores from Saliency Map Clusters for Improved Image-based Harvest-Readiness Prediction in Cauliflower}

\author[1]{Jana~Kierdorf}
\author[1,2]{Ribana~Roscher}

\affil[1]{Remote Sensing Group, Institute of Geodesy and Geoinformation, University of Bonn, Germany}
\affil[2]{Data Science for Crop Systems, Institute of Bio- and Geosciences, IBG-2: Plant Sciences, Forschungszentrum Jülich GmbH}

\corraddress{Jana Kierdorf, Remote Sensing Group, Institute of Geodesy and Geoinformation, University of Bonn,d Germany}
\corremail{jkierdorf@uni-bonn.de}

\runningauthor{\textbf{Preprint Version} of Reliability Scores from Saliency Map Clusters for Improved Image-based Harvest-Readiness Prediction in Cauliflower}  

\fundinginfo{European Agriculture Fund for Rural Development with contributions from North-Rhine Westphalia (17-02.12.01 - 10/16 – EP- 0004617925-19-001); Deutsche Forschungsgemeinschaft (DFG, German Research Foundation) (RO4839/7-1 | STO 1087/2-1); DFG under Germany’s Excellence Strategy – EXC 2070 – 390732324}

\begin{document}

\begin{frontmatter}
\maketitle

\begin{abstract}
Cauliflower is a hand-harvested crop that must fulfill high-quality standards in sales making the timing of harvest important. 
However, accurately determining harvest-readiness can be challenging due to the cauliflower head being covered by its canopy.
While deep learning enables automated harvest-readiness estimation, errors can occur due to field-variability and limited training data.
In this paper, we analyze the reliability of a harvest-readiness classifier with interpretable machine learning. 
By identifying clusters of saliency maps, we derive reliability scores for each classification result using knowledge about the domain and the image properties.
For unseen data, the reliability can be used to (i) inform farmers to improve their decision-making and (ii) increase the model prediction accuracy.
Using RGB images of single cauliflower plants at different developmental stages from the GrowliFlower dataset  
\cite{kierdorf2022growliflower}, we investigate various saliency mapping approaches and find that they result in different quality of reliability scores. With the most suitable interpretation tool, we adjust the classification result and achieve a 15.72\% improvement of the overall accuracy to 88.14\% and a 15.44\% improvement of the average class accuracy to 88.52\% for the GrowliFlower dataset. 


\end{abstract}
\end{frontmatter}

\section{Introduction} \label{sec:introduction}
\label{sec:intro}

Accurate harvest time forecasts are crucial for crop quantity and profitability in agriculture. For cauliflower, high-quality requirements for sale further complicate this process. To meet these standards, harvesting must be precisely timed within a short window. 
Since cauliflower growth is highly affected by climate, fields planted at different times may be ready for harvest simultaneously, and plants may develop differently within a single field. Therefore, it is a common agricultural practice that workers harvest plants individually by hand at different times. As the cauliflower head is covered by its canopy, the workers touch the head inside the plant and estimate the size, making harvesting highly time-consuming. 

In digital agriculture, field monitoring is supported by satellite or UAV imagery \cite{fountas2020future} to observe plant development throughout the entire growth period. Machine learning methods increasingly form the basis for analyzing the acquired data, for example, to classify crop ripeness on a large scale \cite{lary2016machine} or to provide detailed predictions about harvest ripeness, the amount of harvest, or the date of harvest-readiness \cite{van2020crop}. Predicting crop traits related to harvest is of economic benefit to farmers, so the model must be reliable, and the farmer should be able to have confidence in the model's decision. 

We address the task of harvest-readiness estimation of single cauliflower plants and aim to derive a reliability score for the model's output that can be used to support the farmer in their decision-making process.
%
To reach our goal, we use saliency mapping to identify image regions that have distinctive characteristics important for the model decision \cite{roscher2020explain,brahimi2018deep}. We extend the clustering approach of saliency maps by \cite{lapuschkin2019unmasking} and combine the maps with knowledge about our application domain and the image properties to derive reliability scores of the model's output.
Similar to our approach, some previous works also aim to improve the model through the integration of interpretations and explanations \cite{weber2022beyond,schramowski2020making,taejoo2022instanceaware}. However, these works present ad hoc frameworks where the interaction between model and explanation is either learned during training or integrated through human interactions via retraining. Our work differs in that we propose a framework for deriving a reliability score for classification predictions that operates post-hoc during inference time without human interaction. Thus, the system can be applied to already trained models without changing the model architecture and without the need for re-training.

The main contributions of this paper are:
\begin{itemize}
    \item a versatile post-hoc approach to derive intuitive reliability scores without time-consuming human interaction;
    \item a use case where the reliability scores are used to improve harvest-readiness predictions on the GrowliFlower dataset by 15.73\% to an overall accuracy of 88.14\% and by 15.44\% to an average class accuracy of 88.52\%.   
\end{itemize}

\section{Materials and Methods}

\begin{figure*}[!t]
\centering
\includegraphics[width=\textwidth]{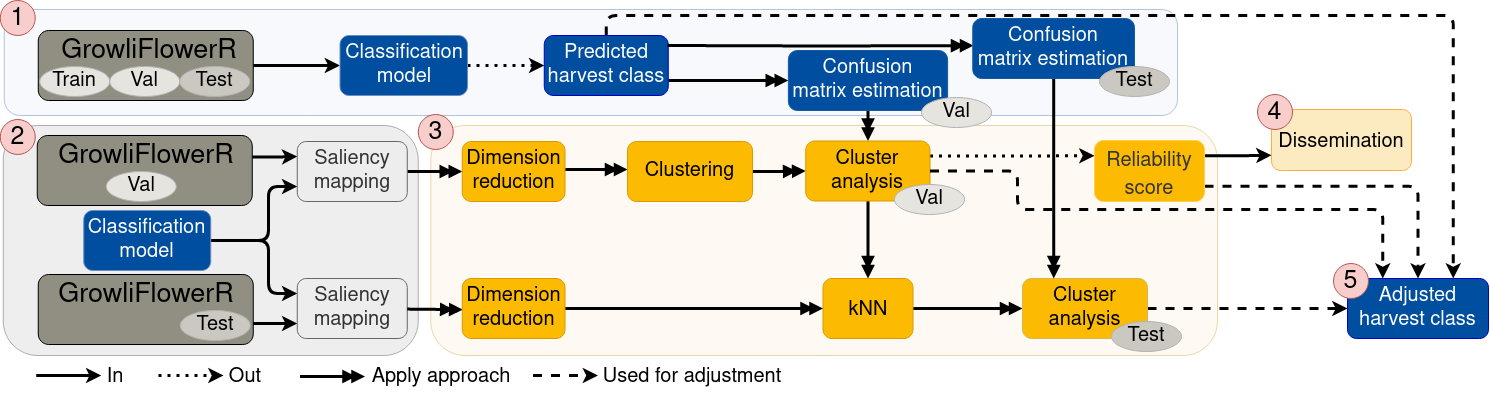}
\caption{Our framework. The different numbers represent (1) the classification step, (2) the saliency mapping step, and (3) the clustering step of saliency maps with the assignment of reliability to the clusters by relating the confidence scores of the model to the corresponding saliency maps. (4) represents the dissemination to the farmer how reliable the model is while (5) represents the adjustment step, where the predictions of (1) are improved by using the reliability score of (3).}
\label{fig:approach}
\end{figure*}

\subsection{Framework}

We solve the task of estimating the harvest-readiness of single cauliflower plants with deep learning-based image classification and combine it with an estimation of the reliability of the classification through clustering of saliency maps. \figref{fig:approach} shows an overview of the five-step framework.

(1) \textit{Classification:}
In the first step, images are classified into the classes \texttt{Ready} and \texttt{Not ready} for harvest within three days. We use a ResNet18 network \cite{he2016deep}, however, the framework is flexible regarding the classifier.

(2) \textit{Saliency mapping:}
In the second step, we compute saliency maps for validation and test data post-hoc using the learned classifier. We consider Gradient-weighted Class Activation Mapping (Grad-CAM) \cite{selvaraju2017grad}, Occlusion Sensitivity Mapping (OSM) \cite{zeiler2014visualizing}, and Local Interpretable Model-Agnostic Explanations (LIME) \cite{ribeiro2016should}.

(3) \textit{Clustering:}
We employ spectral clustering to identify groups of saliency maps computed on the validation data and derive reliability scores. The mean saliency map per cluster, denoted as prototype, is further analyzed. Test data can be assigned a reliability score by assigning its saliency map to the nearest cluster.

(4) \textit{Dissemination:}
The reliability score is intuitively usable due to its value range between 0 and 1 and is communicated to the user together with the classification result.

(5) \textit{Adjustment:}
In our use case, the classification results are adjusted based on the cluster assignments of the saliency maps to determine the final predicted classes. The decision depends on the summed percentage of false positives (FP) and false negatives (FN) per cluster. The evaluation of the classification step provides the assignments to FP and FN.

The framework does not require human interaction and can be applied to different models. However, human interaction is possible to further improve the classification results and reliability measures by analyzing and evaluating the human-understandable clusters of saliency maps.

\subsection{Data}

We use the images of the GrowliFlowerR dataset \cite{kierdorf2022growliflower} of Field~2 from the dates $2021\_08\_23$, $2021\_08\_25$, $2021\_08\_30$, $2021\_09\_03$ with given information about the harvest-readiness within the next three days. Three days is a compromise between harvest-readiness prediction accuracy and practicability of data acquisition. We divide the data into the classes \texttt{Not ready} and \texttt{Ready}. The plants representing both classes show a high similarity within the same day of acquisition but also between different days. The size of its head determines the ripeness, however, in most images, the canopy covers the head. The plant's stem is centered within the image, but depending on the plant's growth, the center of the cauliflower head can vary up to \SI{20}{cm} from the stem.

We use the training, validation, and test set as described in \cite{kierdorf2022growliflower}. If the plant shown in an image is already harvested, we exclude the image from the dataset. This results in a preliminary training set of 541 images, a validation set of 196 images, and a test set of 194 images. We apply standard augmentations like flipping and rotation on the training data. For images of class \texttt{Not ready}, we apply augmentations 50\% more often than for images of class \texttt{Ready} to get a more balanced data distribution. After data augmentation, the training set contains 6224 images, 2432 of class \texttt{Not ready}, and 3792 images of class \texttt{Ready}.

For each image, we compute corresponding saliency maps. The datasets result in pairs of image and map. Thus, all target information of the images are also valid for the corresponding saliency maps.

\subsection{Classification}
\label{sec:classification}

We use a ResNet18 \cite{he2016deep} architecture with cross-entropy loss, softmax activation, and two classes as output. We compute the model over 25 epochs and  use an Adam optimizer with a weight decay of 0.0001. The learning rate starts at 0.0001 and is reduced with a learning rate scheduler with a step size of 5 and factor $\gamma$ of 0.1.

\subsection{Saliency maps}
\label{sec:saliency_maps}
Saliency maps aim to explain a model's decision by identifying important regions in the image. In our case, saliency maps highlight which image regions are important for predicting the classes \texttt{Ready} and \texttt{Not ready}, allowing conclusions about the reliability using the prior knowledge that the center of the image is important for the decision and the background should not play a role in the harvest-readiness estimation.
We consider three local approaches,that represent the basic principles of saliency mapping, namely a gradient-based approach, Grad-CAM, and two permutation-based approaches, OSM and LIME, where LIME differs in that it uses surrogate models, as our focus is not on the used methods.

Grad-CAM is a gradient-based model-specific method developed by Selveraju et~al. \cite{selvaraju2017grad} that uses gradient information to determine from which image regions the convolutional layer takes the information for prediction. The resulting map depends on the employed layer, where we follow the suggestions of Selveraju et~al. \cite{selvaraju2017grad} to use the last convolutional layer as it highlights object-level regions in the image, which are also easier to interpret. Grad-CAM provides information about the class of interest but no information about other classes.

The second approach, OSM, is a perturbation-based model-agnostic method developed by Zeiler et~al. \cite{zeiler2014visualizing}.
This method evaluates sensitivity towards occlusion. It uses a sliding window approach with patchsize $p$ and stride $s$ to permute the input by masking patches and, thus, determine the influence of the occlusion on the predicted model score. 
A blue pixel in the map indicates that the score after occlusion is lower than the original score, i.e., this pixel indicates the presence of the examined class.
A red pixel indicates that the score after occlusion is higher than the original score, indicating a different class.
Note that the smaller $s$, the finer the map's resolution. In our experiments, we chose $s=2$ and $p=11$.

Like OSM, the third approach, LIME, is a perturbation-based model-agnostic method developed by Ribeiro et~al. \cite{ribeiro2016should}. LIME perturbs the input and computes the prediction for these perturbed samples with the original model. Perturbation is applied by changing components in images that are meaningful to humans, such as superpixels. After perturbation, a local surrogate model is learned using the perturbed samples. In our work, we use a least squares linear regression model.

\subsection{Spectral Clustering}
\label{sec:clustering}
We follow the idea of Lapuschkin et~al. \cite{lapuschkin2019unmasking} by using Spectral Clustering (SC) introduced by Ng et~al. \cite{ng2002spectral} to cluster the resulting saliency maps, which provides a better understanding of the model decision by taking into account image features other than RGB.
SC involves clustering data based on a similarity measure derived from a new representation of the data. As similarity, we chose Gaussian similarity function with a kernel scale of 0.2 based on the Euclidean distance. Before we apply SC on our saliency maps, we perform principal component analysis on the vectorized data to reduce the dimensions of the data from 65536 to 50. 
We decided on a dimension of 50 to obtain 95\% of the variance because there is no unique eigenvalue difference, i.e. successive eigenvalues have no significant difference.
We apply SC to the validation set and assign the closest cluster IDs to test data using kNN with $k=5$. For our approach, we set the number of clusters $q=8$ to be representative and generalizable to other datasets.

\subsection{Evaluation metrics}
\label{sec:metrics}

To evaluate the adjustment step, the summed percentage of FP and FN is considered in the calculated clusters $q$. We define this percentage as reliability score $r_q$. The higher the reliability score the more unreliable is a prediction in a specific cluster. If $r_q$ exceeds a threshold $t$ in cluster $q$ of the validation set, we swap the predicted class for all samples in cluster $q$ and update the confusion matrix. We choose $t=75\%$ based on a significant improvement in the validation set's accuracy. Threshold $t$ is variable and selectable based on experience. Based on the updated confusion matrix, we adjust the overall and average class accuracy.
We store the identified clusters for swapping and apply the same to the test data, followed by updating the test confusion matrix and accuracies.

\section{Results and Discussion}
We run our experiments on an AMD EPYC 7742 64-Core processor and an NVIDIA A100 for PCIe graphic card with 40 GB hBM2 RAM.

\subsection{General discussion}
Our experiments find that clusters and harvest-readiness classes do not correlate. This is expected in the case of binary decision-making, where both classes may end up in the same cluster since they ideally use the same features. Instead, we focus on whether data within a cluster are correctly classified or not, which allows conclusions to be drawn about the reliability of the result. We use the confusion matrix for analysis.
To assist the farmer in making harvesting decisions, we exploit the fact that the saliency maps of plant images end up in clusters whose classification result is primarily on the main diagonal of the confusion matrix (TP or TN) and maps that are associated with incorrect classification results (FP or FN) tend to end up in separate clusters.

\subsection{Classification of harvest-readiness}
\label{sec:res_classification}
On the validation set, we achieve an overall accuracy of 76.32\% and an average class accuracy of 77.21\%.
For inference, we achieve an overall accuracy in classification of 72.41\% and an average class accuracy of 73.08\%.
That means we are able to predict the harvest-readiness of approx. 3 out of 4 plants correctly.

\subsection{Local analysis: Saliency maps of single sample inputs}
\label{sec:local_analysis}

\begin{figure}[!t]
\centering
    \includegraphics[width = 0.47\textwidth]{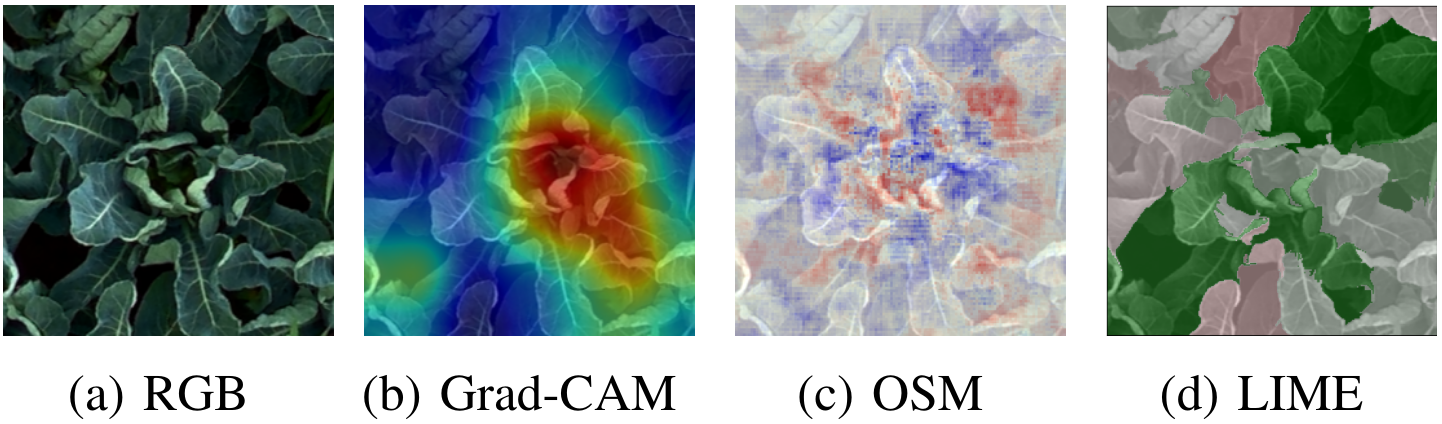}
\caption{Resulting saliency maps for the used approaches (b) Grad-CAM, (c) OSM, and (d) LIME for a RGB input image (a) which is visualized in the maps' background.}
\label{fig:local_saliencymaps}
\end{figure}

In some of the resulting Grad-CAM maps, a hotspot near the center is highlighted in the image as shown in \figref{fig:local_saliencymaps}~b). In other maps, the highlighted regions are located near the edges or scattered in the image. It is easy to analyze which regions have an influence on the model's decisions since compact regions are highlighted.

A considerable amount of the OSM results resemble noise regardless of stride and patchsize for occlusion. Only a minor portion of the results show larger connected regions that are important for decision, as shown in \figref{fig:local_saliencymaps}~c). These are located in the area of the image that shows, among other things, the hidden cauliflower head or highlighted leaf regions. Many maps show several smaller highlighted regions which are difficult to explain because they do not indicate a unique plant trait. The ability of a simple explanation of the results varies more than for Grad-CAM.

In LIME maps, we see that the computed superpixels are not able to summarize pixels to semantically meaningful regions. 
This could be caused by the structure or the strong overlap of neighboring plants. Due to this, LIME saliency maps are difficult to connect to general statements about the reliability of classification outputs. An example of a sample analyzed by LIME is shown in \figref{fig:local_saliencymaps}~d). We consider LIME not suitable for our application.

Based on the assessment of single saliency maps, we consider Grad-CAM and OSM to be the most suitable approaches in our framework.

\subsection{Global analysis: Clustering of saliency maps and reliability derivation}
\label{sec:res_clustering}

\begin{figure}[!t]
\centering
\includegraphics[width=0.47\textwidth]{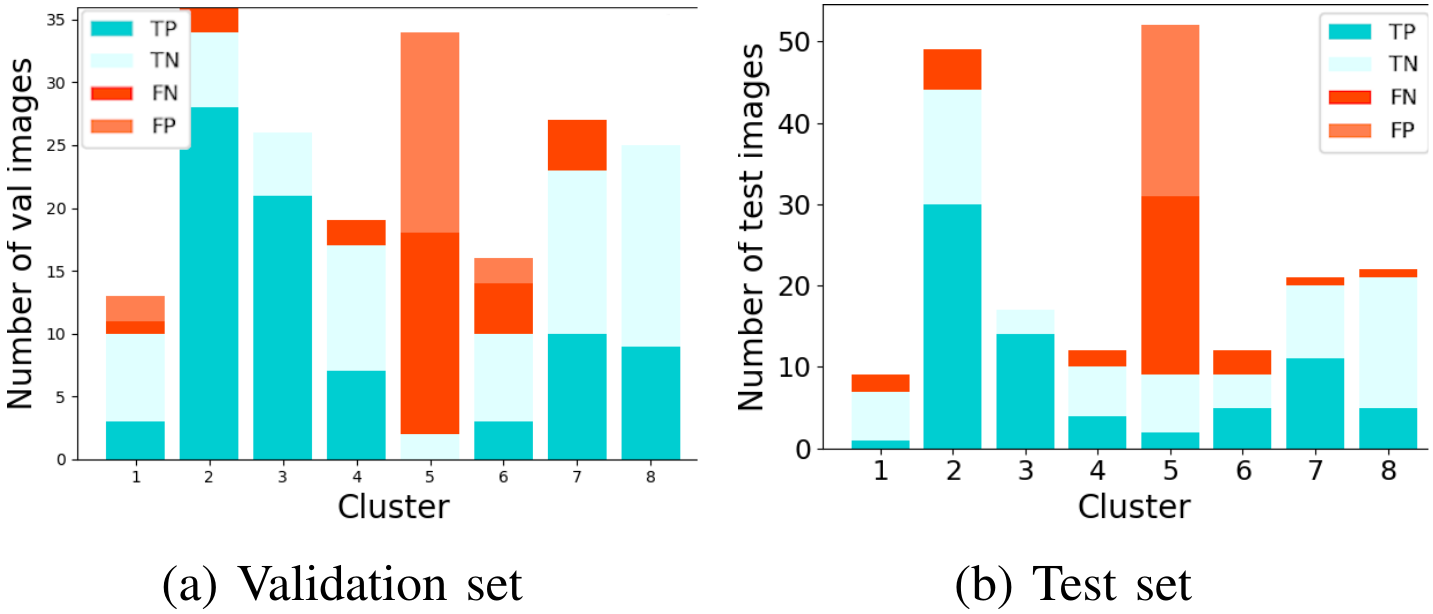}
\caption{Clustering of Grad-CAM results. Absolute number of (a) validation (val) images and (b) test images per cluster.}
\label{fig:res_gradcam}
\end{figure}

\figref{fig:res_gradcam} show the absolute number of Grad-CAM map assignments of the clustering results for 8 clusters. A distinction is made between the validation and test set. The confusion matrix entries are differentiated by color. Our experiments have shown that 8 clusters produce a good separability between false and correct predictions. Furthermore, depending on the amount of data, there are enough data points per cluster to make a reliable statement. 
Based on the distribution of validation data in \figref{fig:res_gradcam}~a), it becomes evident that cluster 5 contains about 95\% false predictions, which are equally divided between FP and FN. This means that over 70\% of all FN and FP belong to cluster 5. The cluster with the second highest proportion of false predictions is cluster 6. It is worth mentioning that the percentage is only 30\%, which corresponds to only six images. 
The other clusters contain less than 20\% false predictions. The clustering analysis allows saying with high confidence that samples assigned to cluster 5 are equivalent to a false prediction and should be adjusted. The reliability of the classification results of the saliency maps assigned to this cluster is, therefore, low and should be disseminated to the farmer.
This is underlined in particular by the cluster assignments of the test data (\figref{fig:res_gradcam}~b). We observe that 80\% of the false predicted test data are assigned to cluster 5. The proportion of false predictions in the other clusters is comparable to those within the validation data.

\begin{figure}[!t]
\centering
\includegraphics[width=0.47\textwidth]{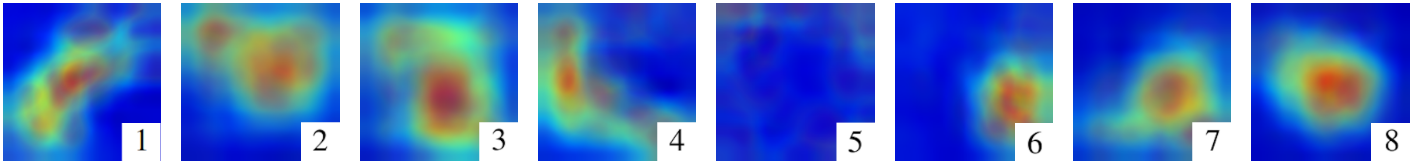}
\caption{Grad-CAM prototypes computed by mean saliency map per cluster (1) -- (8).}
\label{fig:res_mean_gradcam}
\end{figure}

The prototypes of Grad-CAM are shown in \figref{fig:res_mean_gradcam}. Half of the prototypes (2,3,7,8) highlight the region in the center of the image. This is the location in the RGB input images of cauliflower heads covered by leaves, which are the indicators of cauliflower harvest-readiness. Even though the cauliflower head is not directly visible in the images, the model identifies the center of the plant as an essential feature for the classifier to determine the harvest-readiness. The interpretation of the classification results is straightforward and understandable for these clusters. The previously noticed cluster 5 also varies in this representation to the other clusters. In the image data assigned to the cluster, the classification model finds no distinctive features for determining the harvest-readiness. The visualization of the prototypes thus supports the model's reliability in addition to the cluster assignment since the visual representation is easier for the user to understand and interpret.

\begin{figure}[!t]
\centering
\includegraphics[width=0.47\textwidth]{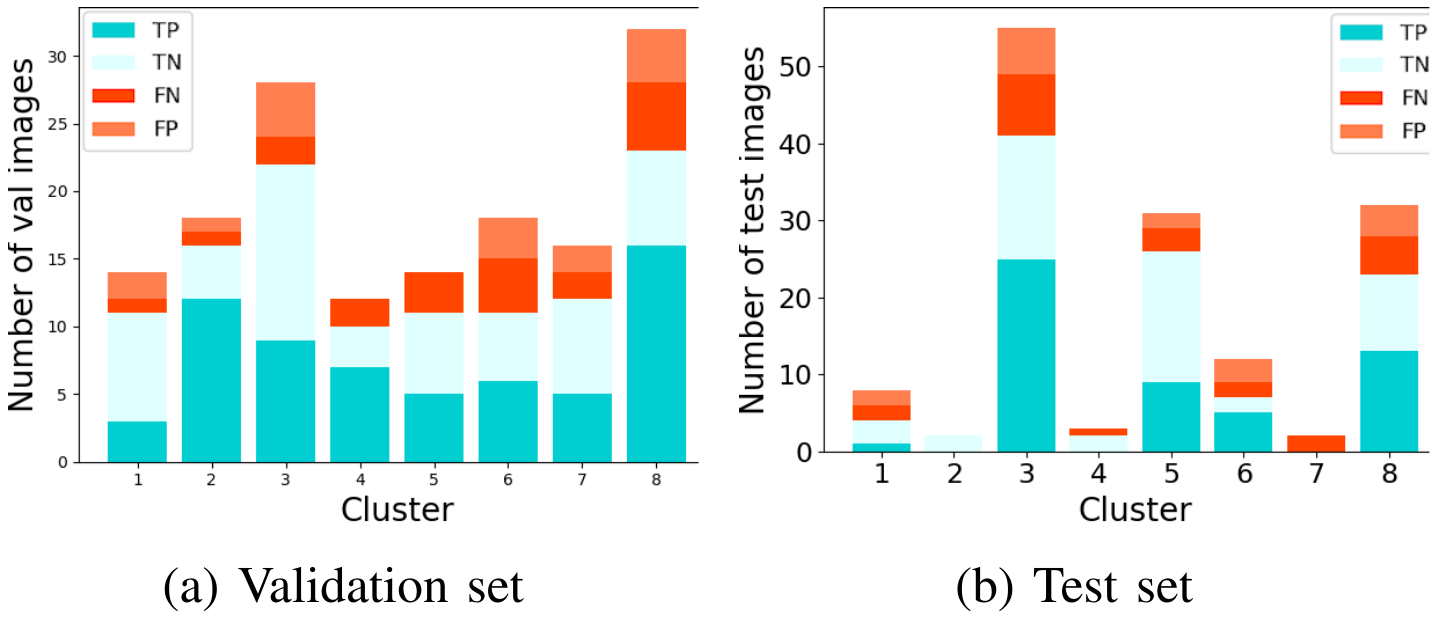}
\caption{Clustering of OSM results. Absolute number of (a) validation (val) images and (b) test images per cluster.}
\label{fig:res_osm}
\end{figure}

The clustering of the OSM maps shows a uniform distribution of false predictions in all clusters (\figref{fig:res_osm}~a). The percentage ranges from 10\% to 30\%. Based on the OSM cluster results, no statement can be made about the reliability of the results. The probability that a false prediction occurs in one of the clusters is similar for all clusters. The cluster assignment of the test data shows a similar distribution (\figref{fig:res_osm}~b). Only cluster 7 stands out. It should be noted that the assignment to this cluster corresponds to a single image only.

\begin{figure}[!t]
\centering
    \includegraphics[width=0.47\textwidth]{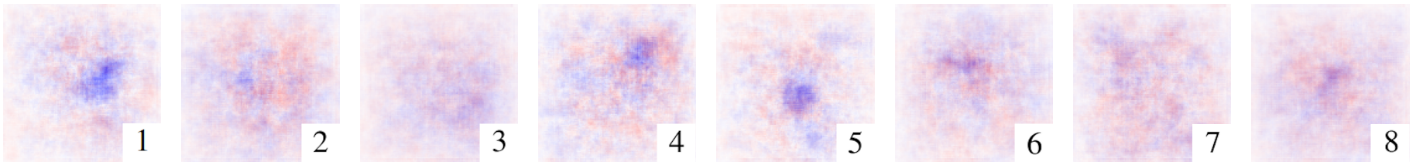}
\caption{OSM prototypes computed by mean saliency map per cluster (1) -- (8).}
\label{fig:res_mean_osm}
\end{figure}

The prototypes also suggest no clear trend in terms of what the model uses as an informative feature in the RGB images (\figref{fig:res_mean_osm}). Clusters 1 and 5 show a hotspot near the center, which, just like Grad-CAM, suggests that the model is paying partial attention to the canopy covering the head. Clusters 4, 6, and 8 give a hint of this.
Comparing the prototypes of the OSM approach with those of the Grad-CAM approach, we see that for our scenario, the Grad-CAM approach results in more interpretable maps than the ones of OSM. Since no clear differentiation between false and correct prediction can be made in the data for OSM, the adjustment step introduced in this work is only applied to the Grad-CAM results. Adjusting the classification results based on the clustering results would worsen rather than improve the model results.

In summary, the combination of saliency map analysis and clustering provides information about the reliability of classification results. Nevertheless, some thought should be given to the saliency mapping approach to be used.

\subsection{Adjustment of model predictions}
With regard to applying the adjustment step to Grad-CAM maps as explained in \secref{sec:metrics}, we achieve a 13.99\% improvement in overall accuracy to 90.31\% and a 13.39\% improvement in average class accuracy to 90.60\% for classification on the validation set. For inference, we achieve a 15.73\% improvement in overall accuracy to 88.14\% and a 15.44\% improvement in average class accuracy to 88.52\%.

\section{Conclusion and Future Directions}
This work proposes a framework to derive a reliability score for cauliflower harvest-readiness estimations that operates post-hoc during inference time without the need for human interaction. Our work combines a ResNet18 classification model with an unsupervised Spectral Clustering approach of saliency maps to derive a reliability score for classification predictions. Since the reliability value is in a fixed range between 0 and 1, it is intuitive and can be provided to the farmer as a decision support. In addition, the classification predictions can be adjusted, and the accuracy can be improved. We compare three saliency mapping approaches: Gradient-weighted Class Activation Mapping, Occlusion Sensitivity Mapping, and Local Interpretable Model-Agnostic Explanations. Grad-CAM proves to be the most useful in our scenario.

For our use case, our approach enables the correct harvest-readiness estimation on GrowliFlowerR, a subset of the GrowliFlower dataset, of approx. 4 out of 5 cauliflowers compared to the state-of-the-art approach ResNet18 which achieves only approx. 3 out of 4 correct predictions.
Our framework offers the advantage of not requiring any interaction with the training process and it can be applied to already trained models without accessing or modifying the model architecture. We provide interpretable visualizations and a reliability score for the model's decision.
Since we only consider false predictions in our framework, the approach can also be used for reliability dissemination in multi-class tasks.

\section*{Acknowledgements}
This project was funded by the European Agriculture Fund for Rural Development with contributions from North-Rhine Westphalia (17-02.12.01 - 10/16 – EP-0004617925-19-001) and 
by the Deutsche Forschungsgemeinschaft (DFG, German Research Foundation) (RO 4839/7-1 $\vert$ STO 1087/2-1).
In addition, this work was supported in part by the DFG under Germany’s Excellence Strategy – EXC 2070 – 390732324.

\appendix

\bibliography{main}

%
%


\end{document}